\documentclass[A4paper, 11 pt, conference]{ieeeconf}  
\IEEEoverridecommandlockouts       
\usepackage{geometry}
\usepackage{graphics} 
\usepackage{epsfig} 
\usepackage{mathptmx} 
\usepackage{times} 
\usepackage{amsmath} 
\usepackage{amssymb}  
\usepackage{url}
\usepackage{subfigure}
\usepackage{algorithm} 
\usepackage{algorithmic} 
\usepackage{multirow} 
\providecommand{\figdir}{pic}
\begin{document}

\title{\LARGE \bf
Booster Lab: A Data-Centric Pipeline for Learning Deployable Humanoid Locomotion Policies
}

\author{Penghui Chen$^{1, 3}$, Tinglong Zheng$^{2, 3}$, Yufeng Zhang$^{3}$, and Mingguo Zhao$^{1, 3}$ 
\thanks{*This research was supported by STI 2030—Major Projects grant number 2021ZD0201402 and Beijing Natural Science Foundation(L243004).}
\thanks{$^{1}$Department of Automation, Tsinghua University, Beijing, China
 }
\thanks{$^{2}$School of Mechanical and Electronic Control Engineering, Beijing Jiaotong University, Beijing, China
 }
\thanks{$^{3}$Booster Robotics Technology Co., Ltd, Beijing, China
 }
}

\maketitle 
\thispagestyle{empty}

\begin{abstract}
Humanoid robot motion learning requires not only task-oriented control policies but also physically feasible and natural behaviors that can be transferred to real robots. However, robot-feasible motion data are often scarce: raw human demonstrations may be incompatible with the robot morphology, open-source clips vary in quality, and simulation-collected robot trajectories still require feasibility checking. To address these challenges, we propose a data-centric training and deployment pipeline that integrates motion data curation, real-to-sim model adaptation, AMP-based reinforcement learning, and sim-to-real deployment. We validate the framework on the Booster T1 robot and further provide preliminary cross-platform validation on Booster K1.
\end{abstract}

\section{Introduction}

Human motion data provide a rich source of prior knowledge for humanoid skill learning. Prior work has used feature-based tracking rewards~\cite{peng2018deepmimic,he2024h2o,fu2024humanplus} and adversarial motion priors~\cite{peng2021amp,escontrela2022adversarial,peng2022ase} to exploit such data for physics-based motion learning. However, raw motion data are rarely ready for robot deployment: human demonstrations can be noisy or inconsistent with the robot morphology, open-source clips vary in quality, and simulation-collected robot trajectories still require feasibility checking and selection.

This makes data processing a central issue for deployable humanoid learning. Retargeting and optimization methods map human motions to robot kinematics while considering contacts, joint limits, and balance constraints~\cite{ayusawa2017motion,darvish2019whole,rouxel2022multicontact}, and recent systems further show the importance of retargeting quality and motion feasibility for downstream policy learning~\cite{araujo2025retargeting,protomotions2025,chen2025ikmr,wang2026spark,zhang2026kdmr}. Yet many pipelines still treat data preparation, policy learning, and real-robot evaluation as separate steps, making deployment failures hard to diagnose and improve.

\begin{figure}[t]
   \centering
   \includegraphics[width=0.9\linewidth]{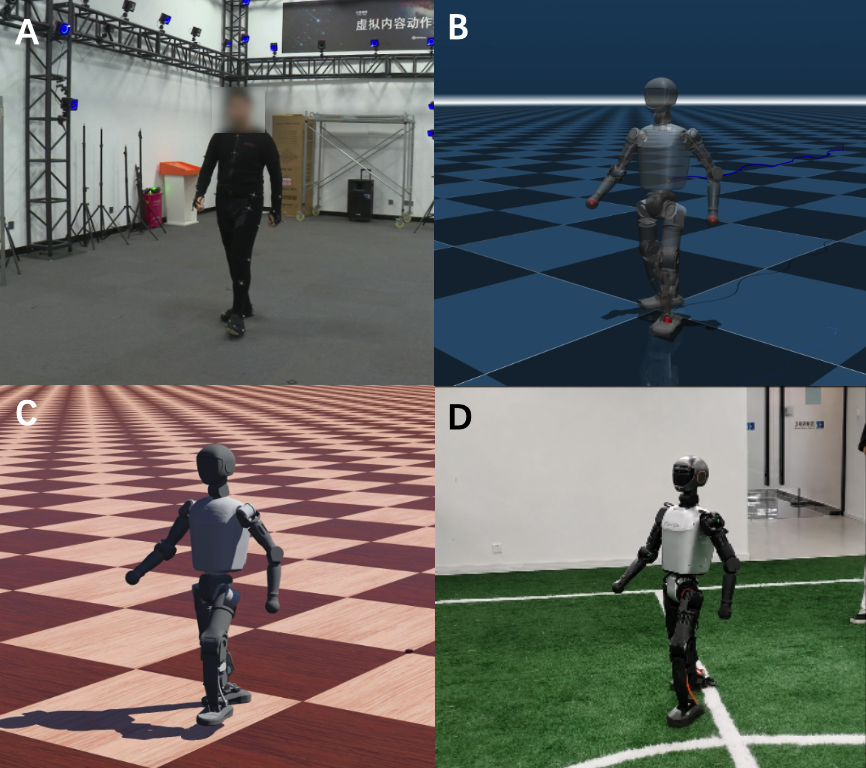}
   \caption{Overview of the proposed pipeline. 
   \textbf{A}: Motion data collection. 
   \textbf{B}: Robot-feasible motion data after curation. 
   \textbf{C}: Policy validation in the adapted simulation model. 
   \textbf{D}: Real-robot deployment and performance evaluation.}
   \label{fig:overview}
\end{figure}

To address this issue, we present \emph{Booster Lab}, a data-centric pipeline for deployable humanoid locomotion learning from heterogeneous motion sources. Motion data are not treated as fixed demonstrations, but as an optimizable component of the learning and deployment loop: their selection, repair, retargeting, augmentation, feasibility filtering, and feedback-driven update are explicitly managed. The pipeline integrates motion/skill selection, motion capture repair, motion retargeting, data augmentation, expert data selection, real-to-sim model adaptation, AMP-based reinforcement learning, cross-simulator validation, and sim-to-real deployment. In this pipeline, real-to-sim adaptation first calibrates the simulation model from real-robot measurements; the adapted model is then used for policy training and validation before the trained policy is transferred to hardware. Real-robot failures can further guide subsequent data selection, augmentation, simulation calibration, and policy updates.

The main contributions of this paper are summarized as follows:
\begin{itemize}
    \item We propose a closed-loop, data-centric pipeline that connects motion data curation, real-to-sim model adaptation, AMP-based reinforcement learning, cross-simulator validation, and real-robot deployment for humanoid locomotion learning.
    \item We construct a robot-compatible expert motion database from heterogeneous motion sources, including human demonstrations, open-source motion clips, and simulation-collected robot trajectories, through repair, retargeting, trajectory-level augmentation, and feasibility-based selection with explicit quality checks.
    \item We validate the proposed framework on the Booster T1 humanoid robot, and further provide preliminary cross-platform validation on Booster K1.
\end{itemize}

\section{Related Work}

\subsection{Learning from Demonstrations}

Learning from demonstrations has been widely used to reduce exploration difficulty and reward engineering in physics-based control. Li et al.~\cite{li2025feature} categorize recent learning-from-demonstrations methods into feature-based approaches, which define explicit tracking rewards from reference motions, and GAN-based approaches, which learn implicit imitation rewards through discriminators.

Feature-based methods usually compare policy states with time-aligned reference states using handcrafted motion features. DeepMimic~\cite{peng2018deepmimic} is a representative example, using dense pose, velocity, end-effector, and root-state rewards to learn physics-based character skills from motion clips. Recent humanoid works extend this direction to whole-body human-to-humanoid imitation and teleoperation, such as H2O~\cite{he2024h2o} and HumanPlus~\cite{fu2024humanplus}. These methods achieve accurate tracking, but often depend on careful retargeting, feature selection, temporal alignment, and reward-weight tuning.

Adversarial imitation methods originate from generative adversarial imitation learning~\cite{ho2016gail}, where a discriminator provides the policy with a learned reward. In motion control, AMP~\cite{peng2021amp} introduced adversarial motion priors to encourage natural motion styles without manually specifying all imitation reward terms. Follow-up works show that AMP-style rewards can replace complex handcrafted rewards~\cite{escontrela2022adversarial} and can be extended with reusable or conditional skill embeddings for diverse motion control~\cite{peng2022ase,tessler2023calm}. Compared with strict frame-level tracking, these methods are better suited to heterogeneous or weakly aligned motion datasets.

Overall, prior work suggests complementary trade-offs between feature-based and adversarial learning from demonstrations in fidelity, interpretability, scalability, and adaptability~\cite{li2025feature}. Our work follows the adversarial motion prior direction, while emphasizing the data pipeline around it: human demonstrations and simulation-collected robot trajectories are curated before being used as expert data for policy training.

\subsection{Motion Data Processing for Robot-Feasible Demonstrations}

Human motion data cannot be directly used as humanoid robot references because of morphology mismatch, joint limits, contacts, and dynamic feasibility constraints. Classical retargeting methods address this problem with geometric morphing, inverse kinematics, and motion optimization. Ayusawa and Yoshida~\cite{ayusawa2017motion} jointly optimized morphing parameters, robot motion, and inverse kinematics to map human motion capture data to humanoid robots. Whole-body geometric retargeting~\cite{darvish2019whole} and multi-contact whole-body optimization~\cite{rouxel2022multicontact} further incorporate robot kinematics, contact constraints, and balance requirements. Recent learning-based humanoid imitation systems such as PHC~\cite{luo2023phc}, H2O~\cite{he2024h2o}, HumanPlus~\cite{fu2024humanplus}, and TWIST~\cite{ze2025twist} also rely on motion retargeting and post-processing to convert human pose data into robot-trackable references.

More recent work studies retargeting quality as a bottleneck for downstream policy learning. GMR~\cite{araujo2025retargeting} provides a general retargeting pipeline for diverse humanoid robots, with practical support for reducing artifacts such as foot sliding, self-penetration, and infeasible contacts. ProtoMotions uses trajectory-level optimization tools for retargeting large-scale SMPL/AMASS motions to robot morphologies~\cite{protomotions2025}. IKMR~\cite{chen2025ikmr}, SPARK~\cite{wang2026spark}, and KDMR~\cite{zhang2026kdmr} further move beyond frame-wise kinematic matching by incorporating kinodynamic refinement, whole-body trajectory optimization, contact consistency, or imitation feedback. These works motivate our data curation stage, where raw demonstrations are repaired, retargeted, checked for feasibility, and selected before being used as expert motion data.

\subsection{Reinforcement Learning Frameworks for Humanoid Skill Learning}

Reinforcement learning has become a dominant approach for humanoid locomotion and skill learning, supported by simulators that can generate large-scale interaction data before real-robot deployment. Isaac Gym~\cite{makoviychuk2021isaac} enabled massively parallel robot learning through GPU-based physics simulation and direct tensor integration. Based on this infrastructure, Humanoid-Gym~\cite{gu2024humanoidgym} provides an Isaac-Gym-based framework for humanoid locomotion with zero-shot sim-to-real transfer and sim-to-sim validation, while Booster Gym~\cite{wang2025boostergym} packages training, reward design, domain randomization, and deployment for the Booster T1 platform.

Recent frameworks further move from task-specific gym environments toward modular robot-learning platforms. Isaac Lab~\cite{mittal2025isaaclab}, the successor of Isaac Gym, unifies GPU-accelerated physics, rendering, actuator models, sensor simulation, data collection, domain randomization, and reinforcement/imitation learning workflows. TienKung-Lab~\cite{tienkunglab2025} follows this trend for full-sized humanoid locomotion with an Isaac-Lab workflow, AMP-style rewards, periodic gait rewards, ray-casting sensors, and sim-to-sim transfer. These frameworks highlight the importance of end-to-end toolchains for simulation, policy optimization, validation, and deployment; our framework focuses on the complementary problem of closing the loop between curated motion data, AMP-based training, and real-robot feedback.

\section{Method}

\begin{figure*}[htbp]
\centering
\includegraphics[width=0.9\linewidth]{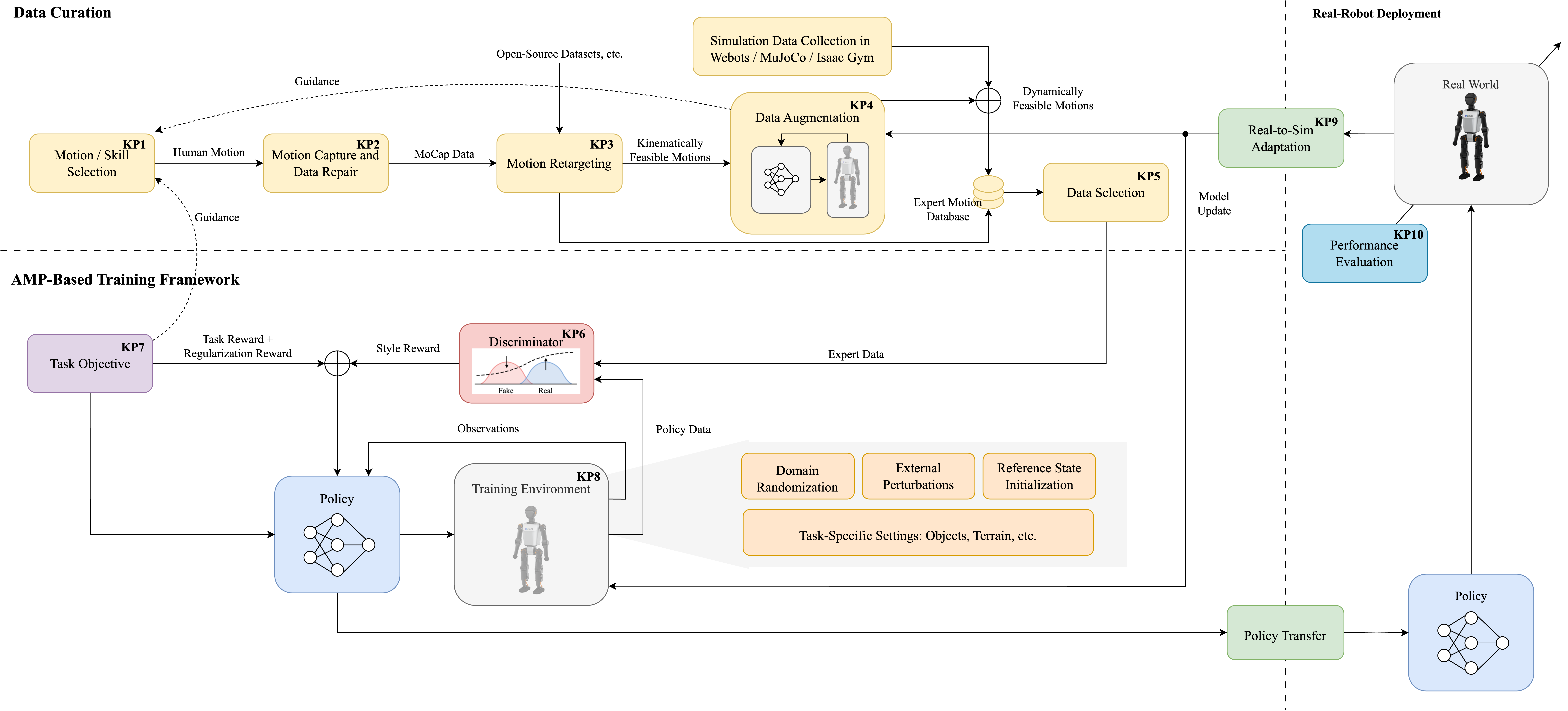}
\caption{The proposed data-centric training and deployment pipeline for humanoid motion learning from heterogeneous motion sources, including human demonstrations and simulation-collected robot data.}
\label{fig:framework}
\end{figure*}

The proposed framework transforms heterogeneous motion sources into deployable Booster T1 locomotion policies. As shown in Fig.~\ref{fig:framework}, the pipeline consists of four coupled stages: motion data curation, real-to-sim model adaptation, AMP-based reinforcement learning, and cross-simulator validation before real-robot deployment. 

\subsection{Problem Formulation}

We consider the problem of learning a velocity-conditioned humanoid locomotion policy from curated robot-compatible motion data. The goal is to track velocity commands while producing natural and physically feasible locomotion behaviors.

Let $\mathcal{D}_{e}$ denote the expert motion database used by AMP. Each motion clip is stored as a JSON trajectory with a frame duration, motion weight, and a sequence of frames
\begin{equation}
    \tau_i = \{x_t, R_t, q_t, p^{ee}_t, v_t, \omega_t, \dot{q}_t, \dot{p}^{ee}_t\}_{t=1}^{T_i},
    \label{eq:motion_frame}
\end{equation}
where $x_t\in\mathbb{R}^{3}$ and $R_t\in\mathbb{R}^{4}$ are the root position and quaternion, $q_t,\dot{q}_t\in\mathbb{R}^{21}$ are the positions and velocities of the 21 controlled joints used by the locomotion policy, $p^{ee}_t,\dot{p}^{ee}_t\in\mathbb{R}^{12}$ are local end-effector positions and velocities for two feet and two hands, and $v_t,\omega_t\in\mathbb{R}^{3}$ are root linear and angular velocities. The policy maps a history of proprioceptive observations to 21 joint-position target offsets:
\begin{equation}
    a_t = \pi_{\theta}(o_{t-9:t}), \quad
    q^{des}_t = q^{default} + 0.25 a_t ,
\end{equation}
where $q^{default}$ is the default standing posture. The controller runs with a physics step of $0.005$~s and a decimation of $4$, giving a policy control period of $0.02$~s.

\subsection{Data Curation}

Data curation turns heterogeneous motion sources into robot-compatible expert clips through five key processes in Fig.~\ref{fig:framework}. The first two processes organize and clean raw motion data, while the latter processes make the motions compatible with the Booster T1 morphology, dynamics, and policy-learning objective.

\begin{table}[htbp]
\centering
\caption{Key Processes in Data Curation.}
\label{tab:data_curation}
\scriptsize
\setlength{\tabcolsep}{2.4pt}
\renewcommand{\arraystretch}{1.06}
\begin{tabular}{p{0.12\linewidth} p{0.31\linewidth} p{0.47\linewidth}}
\hline
\textbf{Step} & \textbf{Process} & \textbf{Function} \\
\hline
KP1 & Motion/skill selection & Select target locomotion skills and useful data sources, including human demonstrations, open-source motions, and simulation-collected robot trajectories. \\
KP2 & Motion capture and repair & Remove corrupted segments, align root heading, smooth noisy signals, and check foot-contact consistency. \\
KP3 & Motion retargeting & Map repaired human motions to Booster T1 using GMR~\cite{araujo2025retargeting}, producing kinematically feasible robot motions. \\
KP4 & Data augmentation & Expand the expert set through trajectory tracking and motion augmentation, including BeyondMimic-style tracking~\cite{liao2025beyondmimic}. \\
KP5 & Data selection & Filter expert clips according to tracking stability, contact consistency, joint and torque limits, foot slip, motion diversity, and task coverage. \\
\hline
\end{tabular}
\end{table}

During data augmentation, we apply trajectory tracking to generate additional robot-feasible expert clips. We use BeyondMimic-style tracking to train a policy on the retargeted trajectories in simulation, which corrects minor kinematic inconsistencies, improves dynamic consistency, and produces motion variations under post-training disturbances. Stable generated trajectories are then passed to the selection stage before being used as expert data for AMP-based policy training.

Simulation-collected trajectories from Webots, MuJoCo, or Isaac Gym are used as candidate robot motion clips. Only stable rollouts that satisfy contact-consistency, joint-limit, torque-limit, foot-slip, and tracking-stability criteria are treated as dynamically feasible robot motions and merged with retargeted human motions. After selection and augmentation, all clips are converted into the AMP JSON format in Eq.~\eqref{eq:motion_frame} and stored in the expert motion database.

Expert clips are selected according to both feasibility and usefulness for AMP-based training. We remove clips with unstable tracking, abnormal root motion, excessive joint-limit violations, inconsistent foot contacts, severe foot sliding, or frequent torque saturation. The selected database is further balanced according to command coverage, gait diversity, and motion smoothness. 

\subsection{Real-to-Sim Model Adaptation}

\begin{figure}[h]
   \centering
   \includegraphics[width=\linewidth]{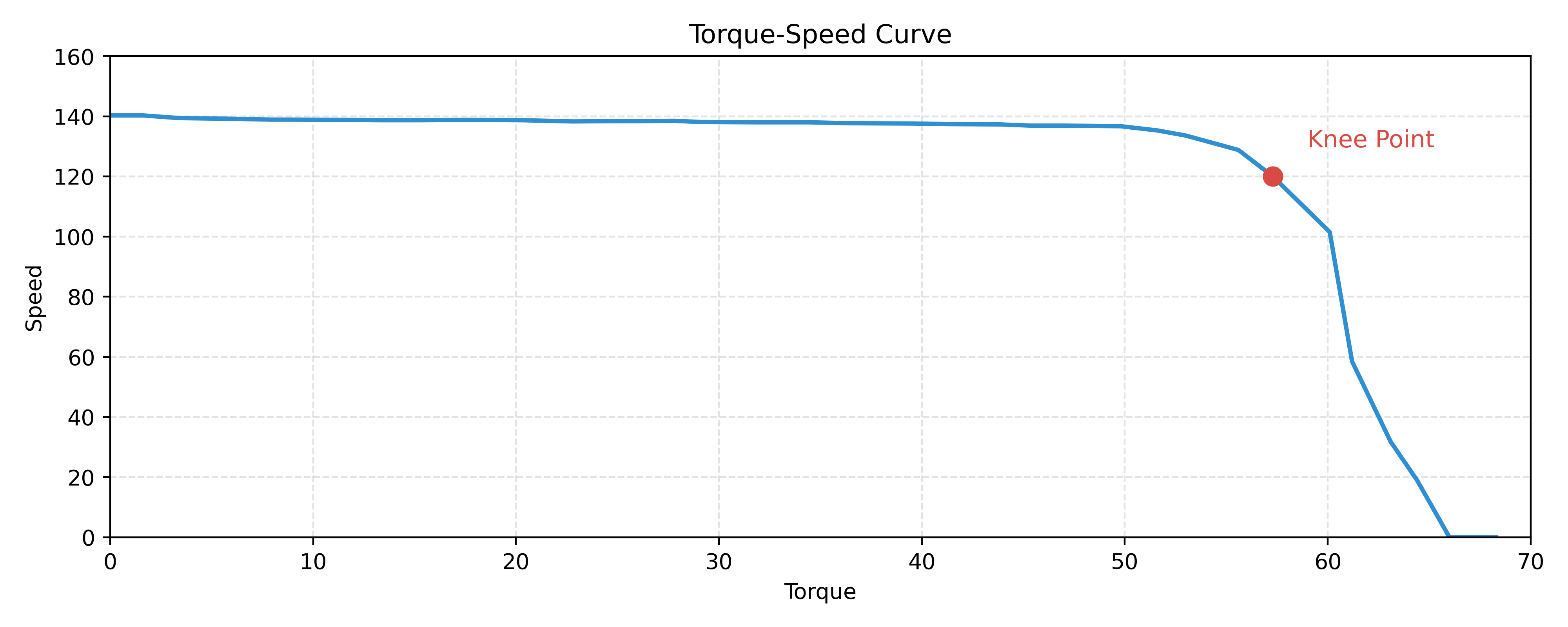}
   \caption{Knee point of the torque--speed \((T\text{-}n)\) characteristic curve.}
   \label{fig:tn}
\end{figure}

Before policy training, we perform real-to-sim adaptation to obtain a simulation model that better matches the Booster T1 hardware. We measure the torque--speed \((T\text{-}n)\) characteristic curves of the Booster T1 actuators and use the knee points of the curves as motor velocity-limit and torque-limit parameters in the simulation model, as shown in Fig.~\ref{fig:tn}. This step adapts the simulator toward the real actuator characteristics and reduces the actuation mismatch faced during later training, validation, and sim-to-real deployment. The resulting joint torque and velocity limits are listed in Table~\ref{tab:tn}.

\begin{table}[htbp]
\centering
\caption{Parameters of Joint Torque and Velocity Limits.}
\label{tab:tn}
\scriptsize
\setlength{\tabcolsep}{2.2pt}
\renewcommand{\arraystretch}{0.96}
\begin{tabular}{p{0.25\linewidth} p{0.31\linewidth} p{0.31\linewidth}}
\hline
\textbf{Joint} & \textbf{Torque limit (N\,m)} & \textbf{Velocity limit (rad/s)} \\
\hline
Waist & 25.0 & 12.57 \\
Hip Pitch & 45.0 & 16.76 \\
Hip Roll & 25.0 & 12.57 \\
Hip Yaw & 25.0 & 12.57 \\
Knee & 60.0 & 12.57 \\
\hline
\end{tabular}
\end{table}

Recent studies~\cite{liao2025beyondmimic} reveal that joint armature can significantly affect humanoid policy transfer. We therefore compute the armature values for the Booster T1 joints based on
\begin{equation}
    J_{\mathrm{arm}} = N^2 J_m
\end{equation}
where $N$ is the gear ratio and $J_m$ is the rotor inertia. The resulting armature values are incorporated into the adapted simulation model. The policy is trained and validated using this adapted model.

\subsection{AMP-Based Policy Training}

The policy training stage uses Isaac Lab with $4096$ parallel environments. Each episode lasts up to $20$~s. The actor receives a 10-step history of a 72-dimensional proprioceptive vector:
\begin{equation}
    o_t = [\omega^b_t, g^b_t, c_t, q_t-q^{default}, \dot{q}_t, a_{t-1}],
\end{equation}
where $\omega^b_t$ is base angular velocity in the body frame, $g^b_t$ is projected gravity, and $c_t=[v_x^{cmd},v_y^{cmd},\omega_z^{cmd}]$ is a sampled velocity command. The critic additionally observes base linear velocity and binary foot-contact states. Velocity commands are sampled every 8--12~s from three ranges: standing commands, normal locomotion commands with $v_x\in[-0.6,1.0]$~m/s, $v_y\in[-0.6,0.6]$~m/s, $\omega_z\in[-1.5,1.5]$~rad/s, and fast forward commands with $v_x\in[1.0,2.0]$~m/s.

The task reward is the weighted sum of velocity tracking and regularization terms summarized in Table~\ref{tab:reward}. The discriminator computes the AMP reward, and PPO optimizes a linear blend of the task and AMP rewards.

\begin{table}[t]
\centering
\caption{Summary of Task Reward Function.}
\label{tab:reward}
\scriptsize
\setlength{\tabcolsep}{2.2pt}
\renewcommand{\arraystretch}{0.96}
\begin{tabular}{p{0.34\linewidth} p{0.43\linewidth} r}
\hline
\textbf{Components} & \textbf{Equations} & \textbf{Weights} \\
\hline
Velocity tracking $(x)$
& $\exp(-(v^{cmd}_{x}-\bar{v}_{x})^2/\sigma^2)$
& $3.0$ \\
Velocity tracking $(y)$
& $\exp(-(v^{cmd}_{y}-\bar{v}_{y})^2/\sigma^2)$
& $3.0$ \\
Velocity tracking (yaw)
& $\exp(-(\omega^{cmd}_{z}-\bar{\omega}_{z})^2/\sigma^2)$
& $3.0$ \\
\hline
Linear velocity $(z)$
& $(v^b_z)^2$
& $-1.0$ \\
Angular velocity $(xy)$
& $\|\omega^b_{xy}\|^2$
& $-0.05$ \\
Energy
& $\|\tau \odot \dot{q}\|$
& $-1\times10^{-3}$ \\
Torque
& $\|\tau\|^2$
& $-1\times10^{-4}$ \\
Joint velocity
& $\|\dot{q}\|^2$
& $-1\times10^{-3}$ \\
Joint acceleration
& $\|\ddot{q}\|^2$
& $-2.5\times10^{-7}$ \\
Action rate
& $\|a_t-a_{t-1}\|^2$
& $-0.1$ \\
Body orientation
& $\|g^{trunk}_{xy}\|^2$
& $-2.0$ \\
Joint position limit
& $\mathbf{1}_{q>q_{\max}}+\mathbf{1}_{q<q_{\min}}$
& $-20.0$ \\
Termination
& $\mathbf{1}_{\mathrm{terminated}}$
& $-200.0$ \\
\hline
Undesired contacts
& $\sum_i \mathbf{1}_{\|f_i\|>1.0}$
& $-1.0$ \\
Feet slide
& $\sum_i \mathbf{1}_{c_i}\|v^{foot}_{i,xy}\|$
& $-0.25$ \\
Feet force
& $\mathrm{clip}(\|f^z_{\mathrm{feet}}\|-500,0,400)$
& $-3\times10^{-3}$ \\
Feet too near
& $\max(0,0.2-\|p^{lf}-p^{rf}\|)$
& $-2.0$ \\
Feet stumble
& $\mathbf{1}_{\|f^{xy}_{\mathrm{feet}}\|>5|f^{z}_{\mathrm{feet}}|}$
& $-2.0$ \\
Feet lateral distance
& $\mathbf{1}_{|v_y^{cmd}|<0.1}||p^{lf}_y-p^{rf}_y|-0.2|$
& $-2.0$ \\
\hline
\end{tabular}
\end{table}

For AMP, the discriminator input is a short history of motion features. At each step, the feature vector contains joint positions, local end-effector positions, root velocities, and joint velocities. Expert transitions are preloaded from $\mathcal{D}_{e}$, while policy transitions are inserted into a replay buffer during rollouts. We use a Wasserstein-style discriminator\cite{gulrajani2017improved} trained to assign higher scores to expert motion snippets than to policy-generated snippets:
\begin{equation}
\begin{aligned}
    \mathcal{L}_{D} =
    &-\mathbb{E}_{z \sim \mathcal{D}_{e}}[D_{\psi}(z)] \\
    &+\mathbb{E}_{z \sim \pi_{\theta}}[D_{\psi}(z)]
    + \lambda_{\mathrm{gp}}\mathcal{L}_{\mathrm{gp}},
\end{aligned}
\end{equation}
where $\mathcal{L}_{\mathrm{gp}}$ is a gradient penalty computed on interpolated expert-policy samples. The AMP reward is derived from the discriminator score:
\begin{equation}
    r^{\mathrm{amp}}_t = \alpha\left(1+\tanh\left(\beta D_{\psi}(z_t)\right)\right), 
\end{equation}
where $\alpha$ and $\beta$ are scaling constants. The final reward used by PPO is a linear blend of AMP and task rewards.

To improve robustness before deployment, we apply domain randomization and external perturbations during policy training. Domain randomization exposes the policy to model mismatch in contact, actuation, body dynamics, initialization, and terrain, as summarized in Table~\ref{tab:domain_randomization}. These terms cover key sources of sim-to-real gap, such as foot-ground contact properties, joint actuation dynamics, hardware inaccuracies, and environment variations.

External perturbations are applied as intermittent horizontal velocity impulses to the robot base, forcing the controller to recover from pushes while maintaining the commanded locomotion behavior. 

\begin{table}[htbp]
\centering
\caption{Domain Randomization Used During Training.}
\label{tab:domain_randomization}
\scriptsize
\setlength{\tabcolsep}{3pt}
\renewcommand{\arraystretch}{1.05}
\begin{tabular}{p{0.34\linewidth} p{0.56\linewidth}}
\hline
\textbf{Category} & \textbf{Randomized Factors} \\
\hline
Contact & Ground friction \\
Actuation & Joint stiffness and damping gains \\
Body dynamics & Trunk mass \\
Initial state & Base pose, base velocity, and joint configuration \\
Terrain & Heightfield roughness \\
\hline
\end{tabular}
\end{table}

Reference state initialization is used as an AMP-specific mechanism rather than a sim-to-real randomization term. At reset, most training episodes are initialized from randomly sampled expert frames instead of the nominal standing pose. The sampled root state and joint state are written into simulation before rollout, allowing the policy to explore around demonstrated motion states from the beginning of an episode. This also provides the discriminator with policy samples closer to the support of the expert motion distribution, which stabilizes adversarial motion-prior learning.

\subsection{Policy Transfer and Cross-Simulator Validation}

After training with the adapted simulation model, the policy is exported for deployment validation. 

We first replay the learned controller in MuJoCo for sim-to-sim validation under an independent dynamics engine. As a final cross-check before hardware deployment, we further run the exported policy in Webots using the deployment-side robot model and interface. Failures observed in MuJoCo, Webots, or real-robot deployment can be used to update the motion set, augmentation choices, randomization ranges, and simulation model settings before retraining.

\section{Experiments}

\subsection{Data Augmentation Ablation}

To analyze the effect of data augmentation, we compare two policies trained with and without augmented expert clips. The non-augmented policy is trained only on retargeted human demonstrations, while the augmented policy is trained on the same retargeted demonstrations after BeyondMimic-style tracking and augmentation. Both policies use the same AMP framework and reward settings.

\begin{figure}[htbp]
   \centering
   \includegraphics[width=\linewidth]{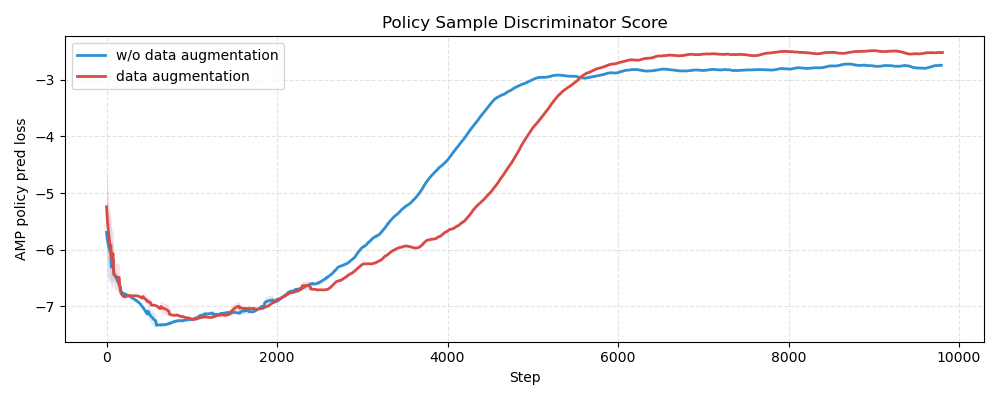}
   \caption{Discriminator predictions on policy samples during training, comparing policies trained with and without data augmentation.}
   \label{fig:augmentation}
\end{figure}

As shown in Fig.~\ref{fig:augmentation}, data augmentation increases the difficulty of AMP imitation in the early training stage, as reflected by lower discriminator predictions on policy samples. As training proceeds, the policy trained with augmented data gradually closes this imitation gap and reaches a higher final policy prediction under the same discriminator setting. This indicates that policy rollouts become closer to the expert distribution under the trained discriminator.

\subsection{Locomotion and Push-recovery Performance}

Booster Gym~\cite{wang2025boostergym} focuses on omnidirectional walking training and deployment for the Booster T1 humanoid robot, our goal is complementary: instead of replacing such gym-style training infrastructure, Booster Lab adds a data-curation and deployment-feedback layer around motion-prior learning. 

We further extend the framework to more dynamic locomotion skills, including walking and running over wider command ranges. The command coverage of the learned policy is shown in Table~\ref{tab:speed}. The policy achieves stable walking from $-0.6$~m/s to $1.0$~m/s and running up to $2.0$~m/s in our hardware trials.

\begin{table}[htbp]
\centering
\caption{Command range of the learned locomotion policy.}
\label{tab:speed}

\begin{tabular}{l c}
\hline
\textbf{Command} & \textbf{Range} \\
\hline
Walking velocity $v_x$ & $[-0.6, 1.0]$~m/s \\
Lateral velocity $v_y$ & $[-0.6, 0.6]$~m/s \\
Running velocity $v_x$ & $(1.0, 2.0]$~m/s \\
Yaw rate $\omega_z$ & $[-1.5, 1.5]$~rad/s \\
\hline
\end{tabular}
\end{table}

\begin{figure}[htbp]
\centering
\includegraphics[width=\linewidth]{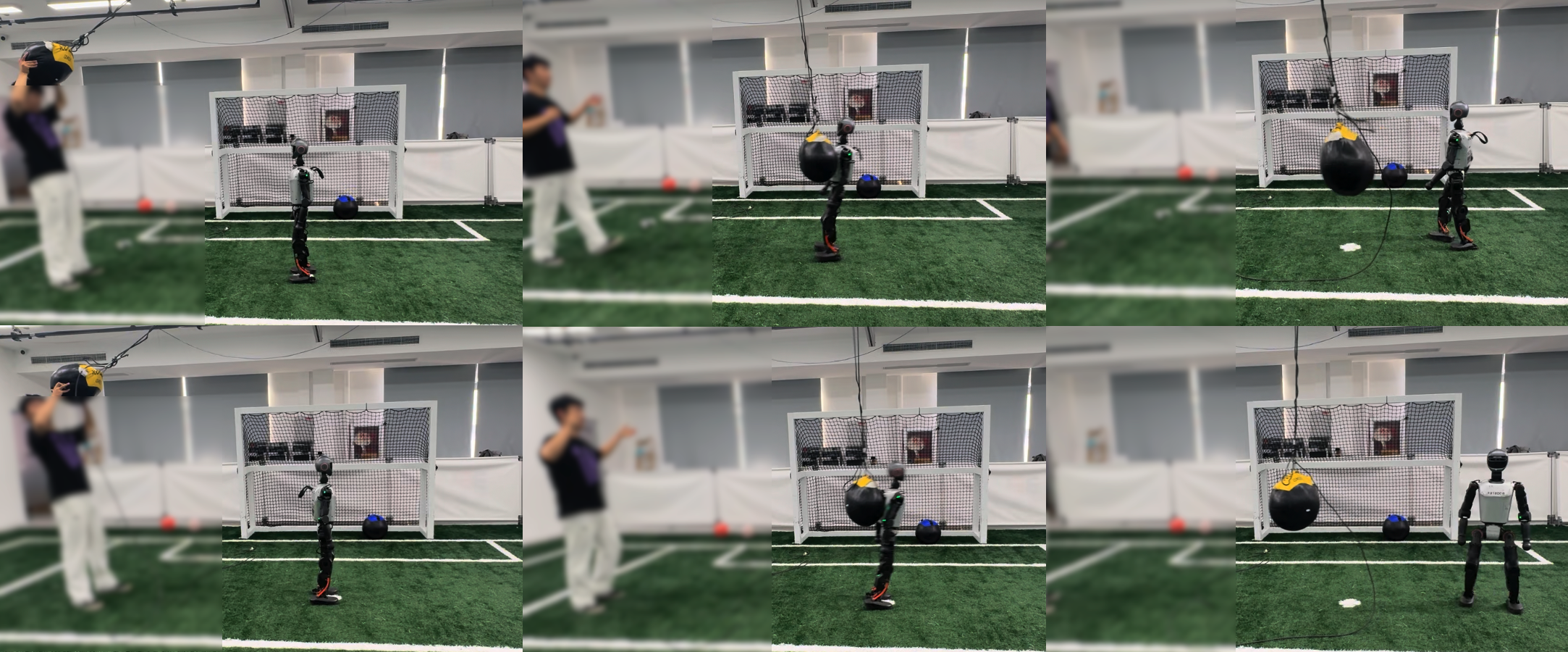}
\caption{Push-recovery performance under external impacts generated by a $10$~kg ball released from a height of $1.3$~m.}
\label{fig:push}
\end{figure}

The learned policy also demonstrates push-recovery capability.
In our hardware tests, external impacts were generated using a $10$~kg ball released from a height of $1.3$~m, with the effective impulse transferred to the robot estimated to be up to approximately $50$~N\,s. Fig.~\ref{fig:push}
shows snapshots of the robot recovering from these disturbances without falling.

\subsection{Locomotion in Outdoor Environments}

\begin{figure}[htbp]
   \centering
   \includegraphics[width=\linewidth]{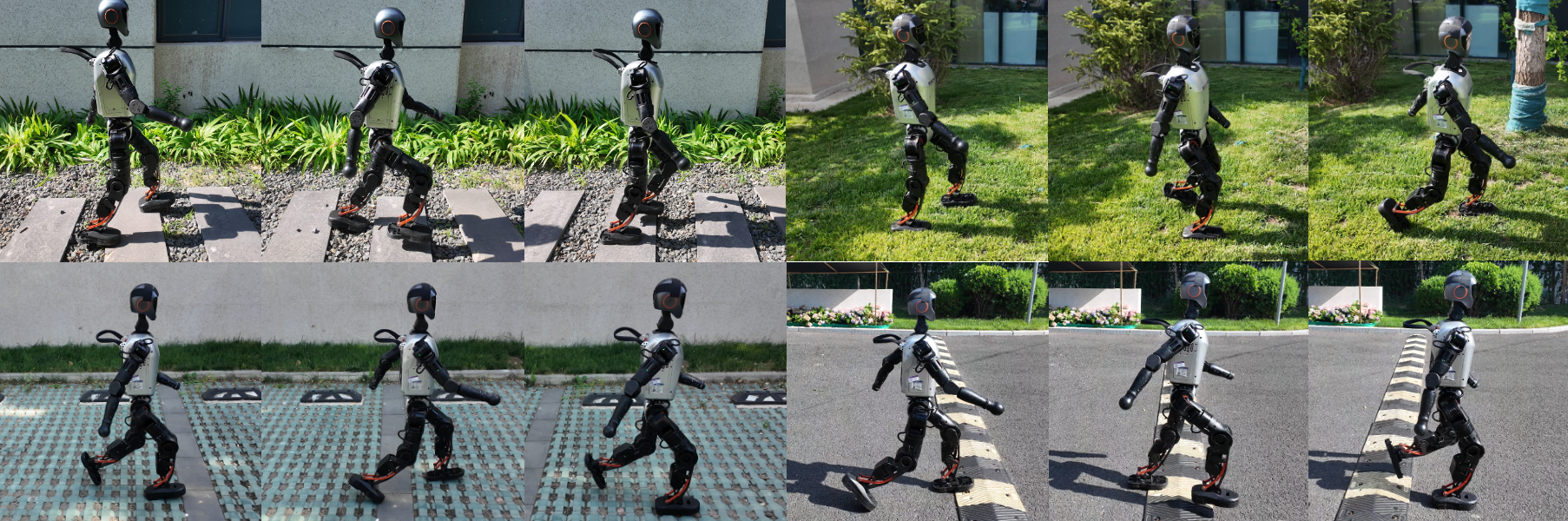}
   \caption{Walking and running in outdoor environments. \textbf{Top left:} running on step stone and gravel. \textbf{Top right:} walking on grass slope. \textbf{Bottom left:} running on uneven terrain. \textbf{Bottom right:} running over the speed bump.}
   \label{fig:outdoor}
\end{figure}

To evaluate the real-world performance of the learned locomotion policy, we deployed it on the Booster T1 humanoid robot and tested it in various outdoor environments. In our hardware tests, the robot walked and ran on different terrains, including step stones, gravel, grass slopes, uneven ground, and speed bumps. Fig.~\ref{fig:outdoor} shows snapshots of the robot executing locomotion skills in these conditions.

\begin{figure*}[t!]
   \centering
   \includegraphics[width=0.9\linewidth]{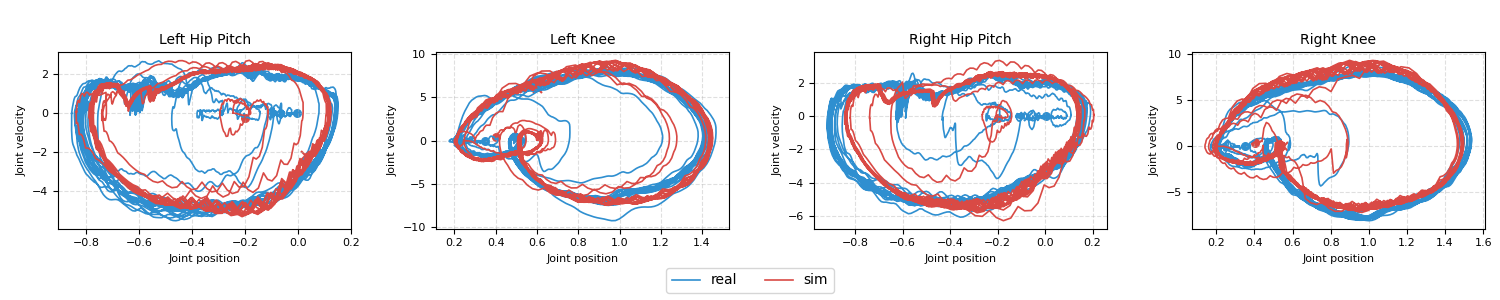}
   \includegraphics[width=0.9\linewidth]{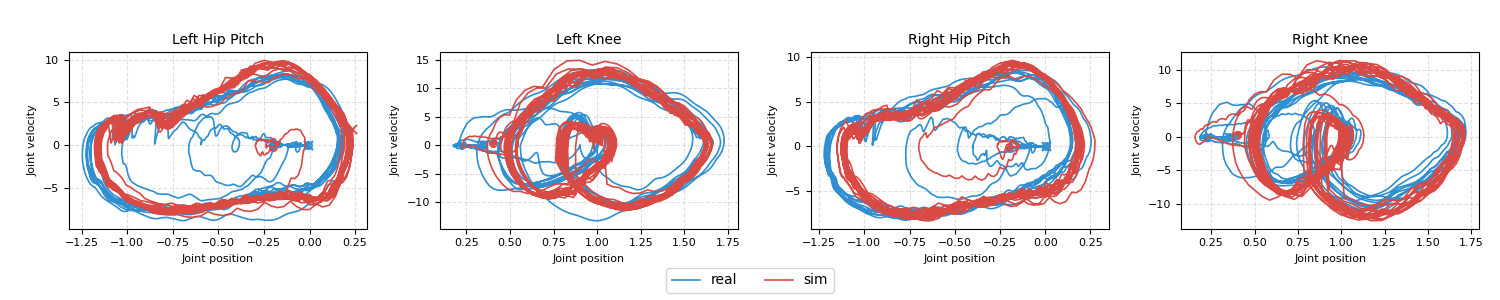}
   \caption{Comparison of hip pitch and knee joint limit cycles between simulation and real-world execution at different speeds. \textbf{Top row:} walking at $0.6$~m/s, \textbf{bottom row:} running at $2.0$~m/s.}
   \label{fig:limit_cycle}
\end{figure*}

With the proposed data-centric pipeline, the learned policy adapts to real-world dynamics and maintains stable locomotion while preserving smooth motion patterns. The successful deployment in outdoor environments further validates the effectiveness of the proposed framework in learning deployable humanoid locomotion skills from heterogeneous motion sources.

\subsection{Sim-to-Real Analysis}

To analyze the sim-to-real transfer of the learned locomotion policies, we compare the limit cycles of key joints during walking and running in both simulation and real-world execution. The limit cycle represents the periodic trajectory of joint angles over a gait cycle, providing insight into the motion patterns and stability of the learned behaviors.

Hip pitch joints and knee joints serve as critical indicators of locomotion quality, as they contribute significantly to stride length, ground clearance, and overall gait dynamics. We extract the joint trajectories from both MuJoCo simulation and real-world trials at different speeds ($0.6$~m/s for walking and $2.0$~m/s for running) and plot the limit cycles in the joint angle space. The results are shown in Fig.~\ref{fig:limit_cycle}.

As shown in Fig.~\ref{fig:limit_cycle}, the simulated and real-world limit cycles have similar shapes. The hip pitch and knee trajectories in simulation closely follow the real-world trajectories, indicating that the learned policy produces consistent motion patterns across both environments, suggesting the usefulness of domain randomization and real-to-sim adaptation in reducing the sim-to-real gap.

\subsection{Cross-Platform Hardware Validation}

To evaluate whether the proposed pipeline is tied to the Booster T1 platform, we further applied Booster Lab to another humanoid robot, Booster K1. Only robot-specific components were replaced, including the morphology model, joint limits, actuator parameters, retargeting configuration, and deployment interface.

\begin{figure}[htbp]
   \centering
   \includegraphics[width=\linewidth]{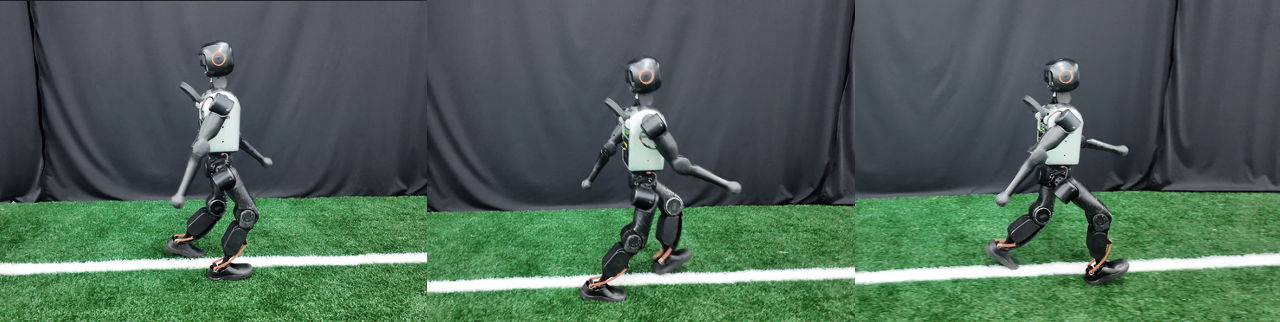}
   \caption{Cross-platform hardware validation on Booster K1.}
   \label{fig:booster_k1}
\end{figure}

The policy achieved stable running on Booster K1, providing preliminary evidence that the proposed data-centric pipeline can be adapted beyond Booster T1. These results suggest that Booster Lab may be transferred across humanoid platforms by updating robot-specific retargeting, actuator modeling, and deployment modules while preserving the same learning and validation workflow.

\section{Conclusion}

In this paper, we presented Booster Lab, a data-centric pipeline for deployable humanoid locomotion learning from heterogeneous motion sources. The framework treats expert motion data as a controllable part of the learning loop and integrates motion data curation, real-to-sim model adaptation, AMP-based reinforcement learning, cross-simulator validation, and sim-to-real deployment.

We validated the proposed pipeline through simulation and real-world experiments on Booster T1, including data-augmentation ablation, push-recovery tests, outdoor locomotion, and sim-to-real limit-cycle analysis. Preliminary validation on Booster K1 further suggests that the same workflow can be adapted to another humanoid platform by updating robot-specific modules.

Future work will extend the framework to more complex skills such as whole-body manipulation, multi-contact locomotion, and dynamic interactions with the environment. We also plan to investigate more advanced data curation and selection strategies to further enhance the quality and diversity of expert motion data for humanoid learning.

\section*{Acknowledgment}
We would like to thank Booster Robotics for providing the
T1 hardware, testing facilities, and technical support.

\bibliographystyle{IEEEtran}
\bibliography{IEEEexample}

\end{document}